\documentclass[lettersize,journal]{IEEEtran}
\usepackage{amsmath,amsfonts}
\usepackage{algorithmic}
\usepackage{algorithm}
\usepackage{array}
\usepackage[caption=false,font=normalsize,labelfont=sf,textfont=sf]{subfig}
\usepackage{textcomp}
\usepackage{stfloats}
\usepackage{url}
\usepackage{verbatim}
\usepackage{graphicx}
\usepackage{cite}
\usepackage{amsmath,amsfonts,amssymb}
\usepackage{multirow} 
\usepackage{multicol} 
\usepackage{booktabs}
\usepackage{utfsym}
\usepackage{arydshln}
\usepackage{color}
\usepackage{makecell}
\begin{document}

\title{Contextual Learning in Fourier Complex Field for\\ VHR Remote Sensing Images}

\author{Yan Zhang, Xiyuan Gao, Qingyan Duan, Jiaxu Leng, Xiao Pu, and Xinbo Gao, \IEEEmembership{Senior Member, IEEE}
\thanks{This work was supported in part by the National Natural Science Foundation of China under Grants 62036007 and 62176195, in part by the Special Project on Technological Innovation and Application Development under Grant cstc2020jscx-dxwtB0032, in part by Chongqing Excellent Scientist Project under Grant cstc2021ycjh-bgzxm0339, and in part by Natural Science Foundation of Chongqing under Grants cstc2021jcyj-msxmX0847. \emph{(Corresponding author: Xinbo Gao.)}}

\thanks{Yan Zhang is with Chongqing Key Laboratory of Image Cognition, Chongqing University of Posts and Telecommunications, Chongqing 400065, China, and also with State Key Laboratory of Integrated Services Networks, Xidian University, Xi’an 710071, China (e-mail: yanzhang1991@cqupt.edu.cn).}
\thanks{Xiyuan Gao, Qingyan Duan, Jiaxu Leng, Xiao Pu, and Xinbo Gao are with Chongqing Key Laboratory of Image Cognition, Chongqing University of Posts and Telecommunications, Chongqing 400065, China (e-mail: S210201025@stu.cqupt.edu.cn; duanqy@cqupt.edu.cn; lengjx@cqupt.edu.cn; puxiao@cqupt.edu.cn; gaoxb@cqupt.edu.cn).}
}

\markboth{Journal of \LaTeX\ Class Files,~Vol.~14, No.~8, August~2021}%
{Shell \MakeLowercase{\textit{et al.}}: A Sample Article Using IEEEtran.cls for IEEE Journals}

\IEEEpubid{0000--0000/00\$00.00~\copyright~2021 IEEE}

\maketitle

\begin{abstract}
Very high-resolution (VHR) remote sensing (RS) image classification is the fundamental task for RS image analysis and understanding. Recently, transformer-based models demonstrated outstanding potential for learning high-order contextual relationships from natural images with general resolution ($\mathbf{\approx224\times224}$ pixels) and achieved remarkable results on general image classification tasks. However, the complexity of the naive transformer grows quadratically with the increase in image size, which prevents transformer-based models from VHR RS image ($\mathbf{\geq500\times500}$ pixels) classification and other computationally expensive downstream tasks. To this end, we propose to decompose the expensive self-attention (SA) into real and imaginary parts via discrete Fourier transform (DFT) and therefore propose an efficient complex self-attention (CSA) mechanism. Benefiting from the conjugated symmetric property of DFT, CSA is capable to model the high-order contextual information with less than half computations of naive SA. To overcome the gradient explosion in Fourier complex field, we replace the \textit{Softmax} function with the carefully designed \textit{Logmax} function to normalize the attention map of CSA and stabilize the gradient propagation. By stacking various layers of CSA blocks, we propose the Fourier Complex Transformer (FCT) model to learn global contextual information from VHR aerial images following the hierarchical manners. Universal experiments conducted on commonly used RS classification data sets demonstrate the effectiveness and efficiency of FCT, especially on very high-resolution RS images. The source code of FCT will be available at \url{https://github.com/Gao-xiyuan/FCT}.
\end{abstract}


\section{Introduction}
\IEEEPARstart{B}{enefited} from the development of aerospace technologies, the availability of VHR remote sensing images have significantly increased and provided plenty of data for analysis. Among the remote sensing (RS) image processing tasks, classification is the fundamental one for RS image understanding and automatic processing. Depending on different classified targets, RS image classification can be divided into image-level classification (i.e., RS image scene classification), pixels-level classification (i.e., RS image semantic segmentation), and mix-level classification (i.e., RS object detection, image matching). With the development of deep learning, RS image analysis and classification has been widely used in many practical applications, such as urban planning\cite{urban}, land use change detection\cite{landusing1,landusing2,landusing3}, environmental monitoring\cite{env-monitor1}, and precision agriculture\cite{precision}. Rather than natural images, RS images are characterized by high spatial, temporal, and spectral resolutions, with various land cover and uncontrollable imaging conditions causing large intra-class variance and small inter-class variance. Therefore, how to efficiently utilize the contextual relationship is the key to addressing VHR RS image classification tasks.
\begin{figure}[t]
    \centering
    \includegraphics[width=\columnwidth]{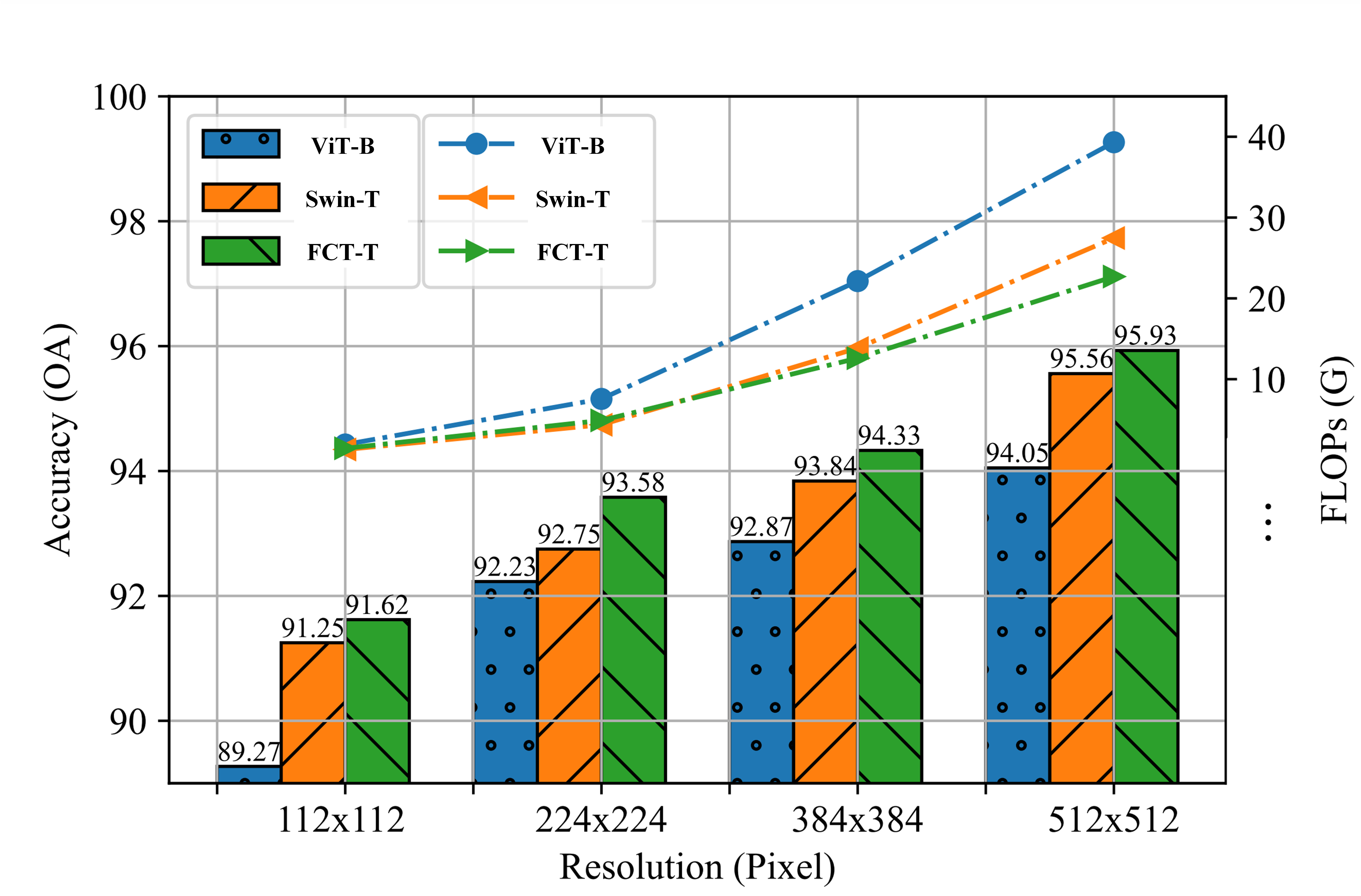}
    \caption{Comparison of the ViT, Swin, and FCT accuracy. $-t$ indicates the tiny version.}
    \label{fig1}
\end{figure}

In the past years, CNN-based models show strong ability and high-efficiency on modeling local information on RS images. Whereas the restriction of the receptive field limited its performance on learning long-range contextual representation. Due to the powerful ability of modeling long-range contextual relationships, Vision Transformer\cite{vit} (ViT) demonstrates promising results and has become the mainstream method on computer vision tasks. Compared with naive CNN, the impressive progress of ViT is mainly thanks to the self-attention (SA) mechanism. \IEEEpubidadjcol SA not only ensures the global receptive field but also learns the high-order features via cross-product operation for better contextual representation. With this, naive ViT surpasses most of the CNN-based methods and achieved state-of-the-art (SOTA) results on the commonly-used ImageNet\cite{imagenet} benchmark. However, there are two common problems need to be overcome when ViT meets VHR images:

\begin{enumerate}[]
    \item The quadratic computational cost ($O(n^2)$) of ViT limits the processed resolution of the input VHR RS image.
    \item Resizing VHR RS image into lower resolution partially relieves problem 1) but causes irretrievably loss of local details and breaks the scale consistency of objects, which seriously affects the performance on RS image classification tasks. 
\end{enumerate}

To improve the efficiency of ViT, several mutations\cite{swin, focal, maxvit} have been proposed. Generally, these methods can be classified into two groups: windowed self-attention and linear self-attention. Swin-transformer (Swin)\cite{swin} is the representative method among the windowed methods. Inspired by the convolution operation, Swin-transformer performs self-attention within a restricted window to ensure the local bias is captured. Also, Swin-transformer proposed a novel shifted window operation to exchange cross-window information. 
As to the Linear self-attention, MaxViT\cite{maxvit} decomposed the pure 2D self-attention into two 1D self-attentions and dramatically reduce the computational costs of self-attention into $O(n)$. Although these mutations show remarkable performances, both of them have significant defects in processing RS images: one of them leads to global information deficiency (windowed SA), and another (Linear SA) is weak in learning abundant 2D contextual representation.


Due to the uniformity of spatial pixel distribution, it is difficult to overcome both of the aforementioned issues in the spatial field. Therefore, we attempted to perform SA in some other fields. Compared with the spatial field, the non-uniformity of pixels in the Fourier complex field provides the possibility to overcome the aforementioned issues in transformer-based models. Also, the orthogonality of the basis functions (sine and cosine) of the Discrete Fourier transform (DFT) ensures transforming pixels from the spatial field into Fourier complex field losses no extra information. For the above 2 reasons, we propose to learn the contextual relationships in the Fourier complex field and further propose an efficiency model named Fourier-based Complex Transformer (FCT) for the VHR RS image classification tasks. The nature of our method is individually performing self-attention mechanism on the sine (imaginary) part and the cosine (real) part to improve the performance of the transformer-based model. To the best of our knowledge, the proposed FCT is the first work to incorporate Discrete Fourier Transform (DFT) into transformer architecture to efficiently learn the contextual relationships (The difference with other Fourier transform-based methods would be compared in Section \uppercase\expandafter{\romannumeral2}. $C$). In FCT, the global representation is obtained through DFT, and the high-order contextual relationships are further learned by the critically designed complex self-attention (CSA) mechanism. As shown in Table \ref{tab_compared}, FCT not only greatly saves computational cost by utilizing the conjugated symmetric property of DFT, but also takes account of high-order relationships with the global receptive field. The contributions of this paper mainly include three aspects as follows:

\begin{enumerate}[]
    \item We analyze the defects of naive self-attention and design an efficient and effective mechanism named Complex self-attention (CSA) to perform contextual relationship learning in the Fourier complex field for VHR RS image classification tasks.
    \item We analyze the difficulty of the gradient backward in the Fourier complex field and therefore design a new normalization function named \textit{Logmax} to stabilize the gradient propagation.
    \item By employing the CSA and \textit{Logmax} function, we design a novel transformer-based network named Fourier-based Complex Transformer, which can learn the global contextual relationship from VHR RS images. With similar computational costs, FCT reports new SOTA results on various VHR RS image classification tasks.
\end{enumerate}

\begin{table}[t]
\caption{Characteristic compared of different methods. $n$ is the size of patches and $m$ is the window size in Swin.}
    \centering
    \begingroup
    \setlength{\tabcolsep}{4pt} 
    \renewcommand{\arraystretch}{1} 
    
    \begin{tabular}{l  c  c  c }
    \toprule
    Method & \makecell[c]{high-order\\ feature} & \makecell[c]{global receptive \\field} & \makecell[c]{computational \\complex}   \\ \midrule \midrule
    CNNs & \usym{2717} & \usym{2717} & $n$   \\
    ViT & \checkmark & \checkmark & $n^2$ \\
    Swin & \checkmark & \usym{2717} & $m^2$\\ 
    GFNet & \usym{2717} &\checkmark	& $nlog(n)$ \\
    FCT & \checkmark & \checkmark & $nlog(n)+\frac{1}{2}n^2$ \\ \bottomrule

    \end{tabular}
    \label{tab_compared}
    \endgroup
\end{table}

\section{Related Work}
\subsection{Remote Sensing Image Analysis}
Remote sensing (RS) images contain plenty of similar and indistinct features and objects, the crucial of RS image analysis is how to extract contextual features efficiently. Feature extraction of RS images can be divided into two eras: the traditionally handcrafted feature-based era and the deep learning-based era. Handcrafted feature-based methods commonly applied manual feature descriptors\cite{hog,sift} and a simple classifier\cite{svm}. Compared with deep learning-based methods, these traditional methods can only extract essentially low-level features, hence the robustness cannot be ensured. Owing to the powerful ability of hierarchical feature extraction, the CNN-based method achieved notable success on various RS image processing tasks. CNNs not only hierarchically extract both the low-level texture and the high-level semantic information but also maintain spatial consistency and therefore producing better results. With fine-tunning trick, Marmanis et al.\cite{pretrained} introduce pre-trained networks learned from large-scale data sets to train the RS image classification model. Henceforth, as the networks have become more complex, it is very popular to apply pre-trained trick to initialize the parameter of networks, i.e., VGG\cite{vgg}, ResNet\cite{resnet}, DenseNet\cite{dense} to tackle RS image tasks. Different from natural images, RS images have significant inner-class variance and confusing information caused by large geographic areas and high-resolution. For this purpose, some researchers have proposed several novel CNN-based methods\cite{abcnet,dualstream,lookingcloser,mgml} specifically designed for RS images, which outperform general CNN-based models by a considerable margin. Nevertheless, the restriction of the receptive field seriously limits the performance of the CNN-based model on VHR RS images. Inspired by the remarkable achievement of the transformer\cite{transformer} in the natural language processing (NLP) domain, Dosovitskiy et al.\cite{vit} first adopted the self-attention mechanism into image processing and proposed the classical Vision Transformer (ViT) to learn contextual relationships for image vision tasks, which surpassed CNNs significantly and achieve promising performance on RS image classification tasks. Nevertheless, the heavy computational costs of ViT are unaffordable when adopting ViT on very high-resolution RS images.

\subsection{Contextual Learning with Transformer}
As analyzed in \cite{swin}, Pure ViT is computationally expensive and lacks some of the inductive biases inherent, which hinder the application on pixel-level and mix-level RS image classification tasks. To relieve these phenomena, various transformer-based mutations\cite{cotnet,mobilevit} are proposed. CoTNet\cite{cotnet} fully utilizes the contextual information among images to guide the learning of attention map and thus enhances the capacity of classification. MobileViT\cite{mobilevit} focused on improving the self-attention mechanism in efficient ways and introduced a new layer that replaced the local convolution processing with a global processing scheme employing self-attention. This strategy afforded a lightweight and low-latency network for mobile vision tasks. In the RS area, Li et al.\cite{grma} suggested a gated recurrent multi-attention neural network, which solved the scattered information problem caused by large regions. Kaselimi et al.\cite{ForestViT} propose a multi-label ViT for the multi-label classification problem in RS image scene. To leverage the advantages of CNN and transformer, Zhang et al.\cite{tandc} generated a hybrid structure to enhance the communication of multi-scale features for RS image segmentation. Generally, although many transformer-based methods have been introduced into RS image classification tasks, the heavy computational costs of SA still prevented the transformer from processing VHR RS images.

\subsection{Neural Network Meets Fourier Transform}
Fourier transform (FT) has been widely used as an essential tool in image processing due to its powerful analytical ability. Fourier transform is a typical mathematical transformation, which decomposes and reconstructs images via trigonometric functions between the real spatial field and Fourier complex field. Moreover, FT is a useful function to aggregate the global information from the real spatial field, where the low-frequency and high-frequency activities reflect the original image's overall character and local variance. With the development of deep neural network (DNN), some researchers attempt to incorporate the FT with DNN. FFC-Net\cite{fourierconv} replaced the convolution in the real spatial field with a local Fourier unit and simulated convolutions in the frequency complex field via a Fast Fourier transform to speed up CNN. To reduce the computational cost and integrate the information of an image token, Lee et al. (FNet)\cite{fnet} replaced self-attention with a simple linear Fourier transformation layer to model diverse relationships in a text. Buchholz et al. (FIT)\cite{fit} designed a Fourier Domain Encoding, which describes the whole image at reduced resolution by using each prefix of the complete image sequence for image super-resolution. Inspired by the frequency filters in digital image processing, Extending from FNet, Rao et al. (GFNet)\cite{gfnet} developed a learnable filter to interchange information in the Fourier field. To optimize partial differential equations in neural networks, Li et al.\cite{fno} constructed a novel neural operator by 
making the integral kernel parameterized in the Fourier field to avoid learning the mapping from any functional parametric reliance to the solution. FrIT\cite{frit} explored a backbone transformer to extract multi-modal global and local contexts from hyper-spectral image (HSI) and LiDAR data with a mix token operation by FT.
Generally, the aforementioned methods mainly utilize FT to speed up convolution or aggregate global features on the Fourier domain. How to learn high-order representation on Fourier complex domain is still lake of concern.


\begin{figure*}[htbp]
    \centering
    \includegraphics[width=0.95\linewidth]{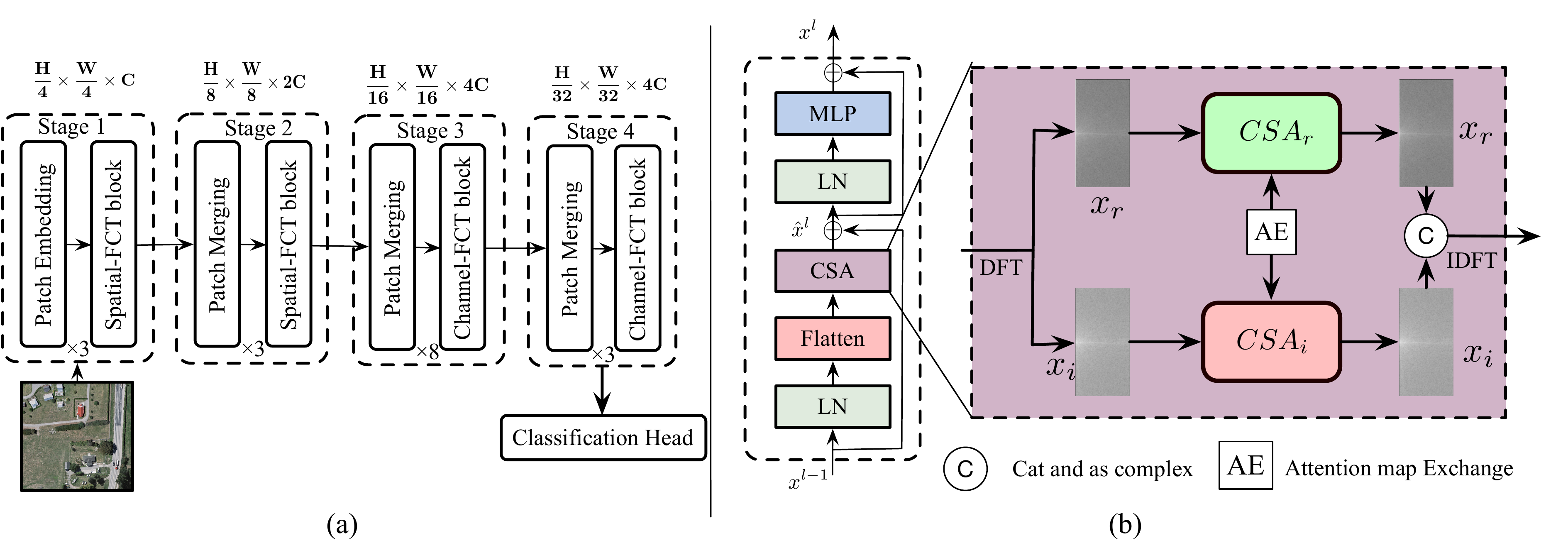}
    \caption{(a) The architecture of the Fourier Complex Transformer (FCT); (b) Fourier Complex Transformer Block.}
    \label{fig_architecture}
\end{figure*}

\section{Proposed Method}
Fig. \ref{fig_architecture}(a) illustrates the overview of the tiny version of Fourier Complex Transformer architecture. FCT is a hierarchical architecture, where its main component is FCT-Block. In each FCT block (Fig. \ref{fig_architecture}(b)), the global representation obtained through the DFT is fed into the carefully designed CSA to learn and extract the contextual relationships. Then, the extracted feature is mapped back into the spatial real field by Inverse Discrete Fourier Transform (IDFT). Finally, we use a learnable linear projection with the Gaussian Error Linear Unit (GELU) activation function and LayerNorm to harmonic the output feature with residual connections. This section is organized as below: Section \uppercase\expandafter{\romannumeral3}.$A$ reviews the preliminaries of Discrete Fourier Transform (DFT) and self-attention mechanism, motivation analysis is also involved in this section. Section \uppercase\expandafter{\romannumeral3}.$B$ elaborates on the details of the proposed FCT block, including \textit{Logmax} and Complex self-attention (CSA). Finally, Section \uppercase\expandafter{\romannumeral3}.$C$ discusses FCT's alternative architecture design and efficiency analysis.

\subsection{Self-attention meets discrete Fourier transform}
Currently, the self-attention mechanism is the most popular method to extract the high-order contextual relationship from images. The self-attention applies the cross product on three learnable vectors $Q, K,$ and $V$. 
Specifically, for input $x$, the result $X$ of self-attention can be reformed as Eq. \eqref{eq1}.
\begin{equation}\label{eq1}
	SA(x) =Softmax\left(Q K^{T}\right) V
\end{equation}
where $Q, K,$ and $V$ are calculated through the linear projection functions $q(x), k(x)$ and $v(x)$, respectively. The critical problem of SA is the quadratic complexity, which prevents SA from the mainstream of processing VHR RS images with limited resources. Hence, projecting the VHR RS image into a simplified space via proper transforms is the fundamental motivation of this paper. 

Discrete Fourier Transform (DFT) is one of the most significant algorithms in digital signal processing, which converts a finite sequence into a same-length complex sequence in the frequency field. Giving a sequence $x$ of length $N$, the DFT transforms $x$ into the Fourier complex field through Eq. \eqref{eq2}.
\begin{equation}\label{eq2}
    \tilde{F}(k)=DFT(n)=\sum_{n=0}^{N-1} x(n) e^{-j(2 \pi / N) k n}
\end{equation}
where $\tilde{F}(k)$ is the complex spectrum of $x(n)$ at the sampling frequency $\omega_k = 2\pi k/N$. Also, the complex frequency feature can be transformed back into the original spatial real field through the Inverse Discrete Fourier Transform:
\begin{equation}\label{eq3}
    x(n)=IDFT(n)=\frac{1}{N} \sum_{n=0}^{N-1} \tilde{F}(k) e^{j(2 \pi / N) k n}
\end{equation}

\begin{figure}[h]
    \centering
    \includegraphics[width=\columnwidth]{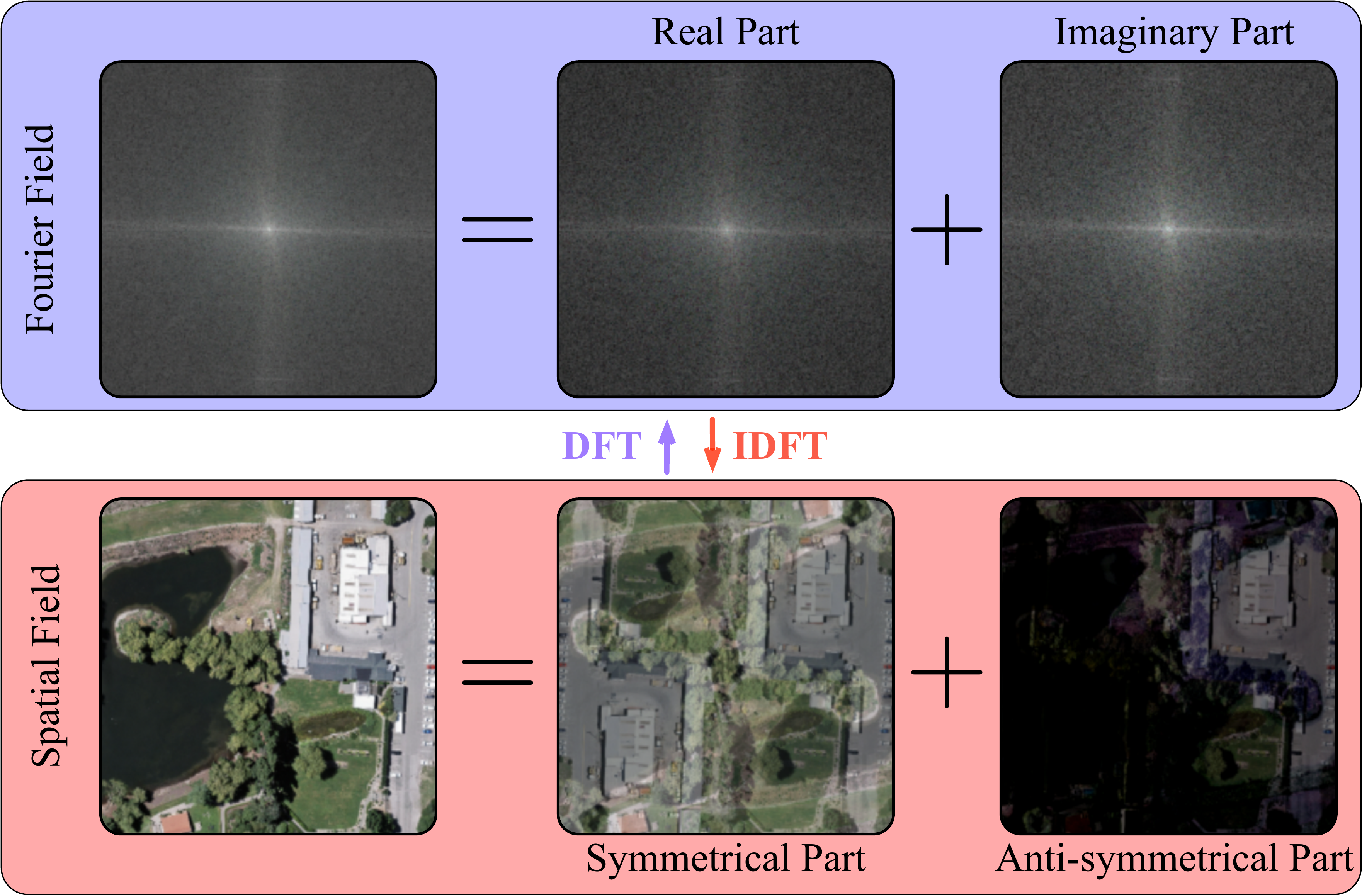}
    \caption{Symmetrical and anti-symmetrical parts from images decomposed via Discrete Fourier Transform.}
    \label{fig_22}
\end{figure}

According to Equation \eqref{eq2}-\eqref{eq3}, DFT and IDFT bridge the spatial real field and frequency complex field. Compared with spatial transforms like affine transformation, one of the appealing feature of DFT is the conjugated symmetric property, where half of the DFT spectrum contains full information of the input in the real field, as well as the efficient algorithm named Fast Discrete Fourier Transform, ensures the computational complexity of DFT and IDFT with $O(nlogn)$. Another helpful feature is that DFT is a global operator to aggregate complex global representation from the real spatial field. The above two features motivate us to improve the design of naive self-attention with the Fourier transform. Therefore, this work utilizes the conjugated symmetric property of DFT to overcome the problem of SA. A feature in the real spatial field of the shape $[H, C]$ can be transformed into the complex field via DFT, while the complex feature's shape can be compressed into $[H, C/2]$. With this simple design, performing SA in the complex field saves nearly three-quarters of the computational burden (from $(HC)^2$ to $\frac{1}{4}(HC)^2$) compared to the real spatial field. Also, the global property of DFT ensures the effectiveness of fully modeling the global information from VHR images. As illustrated in Eq. \eqref{eq4}:

\begin{equation} 
	\label{eq4}
    \left\{
    \begin{aligned}
    X(\omega)&=\int_{-\infty}^{\infty} x(t) \cdot e^{-j \omega t} d t\\&= \int_{-\infty}^{\infty} x(t) \cos \omega t d t - j\int_{-\infty}^{\infty} x(t) \sin \omega t d t\\
    \operatorname{Re}(\omega)&=\int_{-\infty}^{\infty} x(t) \cos \omega t d t\\
    \operatorname{Im}(\omega)&=-\int_{-\infty}^{\infty} x(t) \sin \omega t d t
    \end{aligned}
    \right.
\end{equation}

the naive FT is decomposed into sine part and cosine part via Euler's identity, where the components from the sine part and cosine part are the symmetry and anti-symmetric components of the original image, respectively. The essence of performing SA on the Fourier complex field is individually learning the contextual representations on symmetrical and anti-symmetrical components from a real number image (shown in Fig. \ref{fig_22}).

\subsection{Fourier Complex Transformer}
\subsubsection{Logmax}In this section, we start to analyze the result of directly performing a naive self-attention mechanism in the Fourier complex field. Initially, we simply assume that $Q=K=V=X$ and then we have the naive version of self-attention without \textit{Softmax} function, which can be rewritten as Eq. \eqref{eq5}.
\begin{equation}
\label{eq5}
    DFT(XX^{T} X)=DFT(X)\cdot DFT(X^{T})\cdot DFT(X)
\end{equation}
According to the associative law of DFT, performing self-attention on the Fourier complex field is equivalent to that of the real spatial domain. Therefore, simply and directly performing self-attention without \textit{Softmax} function on the Fourier complex field will gain no extra benefit. Nonetheless, as Eqs. \eqref{eq6}-\eqref{eq7}, performing SA with \textit{Softmax} in Fourier complex field would cause gradient explosion during training, because the differential of \textit{Softmax} function in the Fourier complex field $\frac{\mathrm{d} (Softmax)}{\mathrm{d} (DFT)}$ is unstable, and causing gradient explosion during training. More stabilization analysis of training networks in the Fourier complex field is detailed in the next part.


\begin{flalign}
\label{eq6}
    Softmax(DFT(XX^{T}))DFT(X)&=\\Softmax(DFT(X)DFT(X^{T})) & \nonumber \cdot DFT(X)
\end{flalign}

\begin{equation}\label{eq7}
\begin{aligned}
    \frac{\mathrm{d} (Softmax(DFT))}{\mathrm{d} x} =\frac{\mathrm{d} (Softmax)}{\mathrm{d} (DFT)} \cdot\frac{\mathrm{d} (DFT)}{\mathrm{d} x}
\end{aligned}
\end{equation}

In response to the issue of performing SA in the Fourier complex field, we replace \textit{Softmax} function with a new carefully designed normalized function named \textit{Logmax} function, which effectively stabilizes the gradient with no extra costs. The \textit{Logmax} function is computed as:
\begin{equation}\label{eq8}
    Logmax \left(x_{i}\right)=\frac{\log \left|x_{i}\right|}{\sum_{i} \log \left|x_{i}\right|}
\end{equation}

\subsubsection{Logmax vs Softmax}
\begin{figure}[h]
    \centering
    \includegraphics[width=0.95\columnwidth]{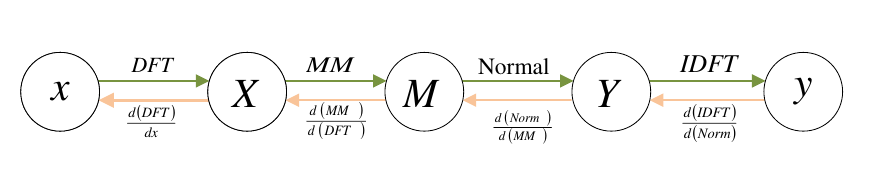}
    \caption{Forward and backward of simplified SA in Fourier complex field.}
    \label{fig_Softmax_vs_Logmax}
\end{figure}

In \textit{Logmax} function, it calculates the probability distribution by $log|x_i|$. While \textit{Logmax} function calculates the probability distribution using $log|x_i|$. To demonstrate the effectiveness of \textit{Logmax} function, we simplified the forward (Eq. \eqref{eq9}) and backward workflows of computing attention map on Fourier complex field in Fig. \ref{fig_Softmax_vs_Logmax}. 

\begin{equation}\label{eq9}
    Y=IDFT(Norm(MM(DFT(x)))) 
\end{equation}

where $x$, $X$, $M$, $Y$ and $y$ indicate the feature in the spatial field, feature map in Fourier complex field, attention map, normalized attention map, and attention map in the spatial field, respectively. $MM$ represents matrix multiplication in Eq .\eqref{eq9}. The gradient during backward propagation is computed as:


\begin{figure}[t]
    \centering
    \includegraphics[width=\columnwidth]{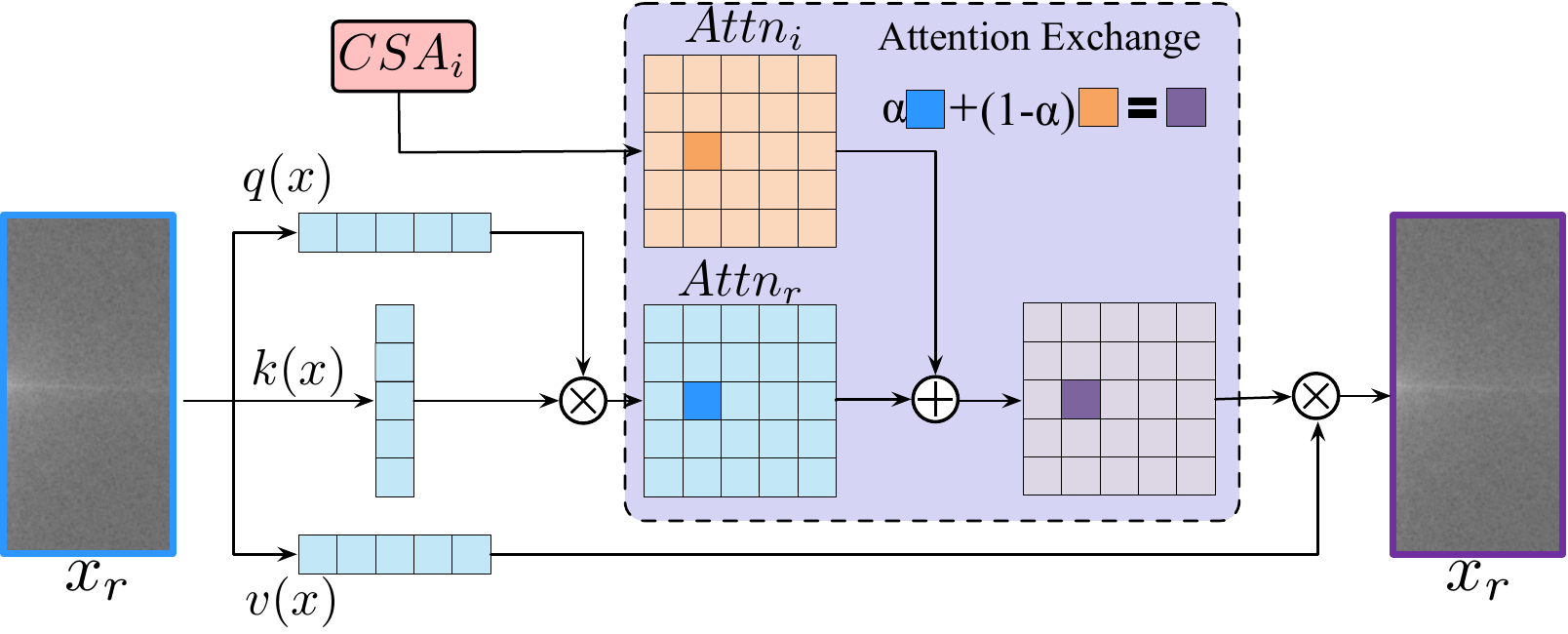}
    \caption{Computational Diagram of Real Part CSA.}
    \label{fig_csa}
\end{figure}

\begin{equation}\label{eq10}
    \frac{\mathrm{d} Y}{\mathrm{d} x} =\frac{\mathrm{d} (IDFT)}{\mathrm{d} (Norm)} \cdot \frac{\mathrm{d} (Norm)}{\mathrm{d} (MM)} \cdot \frac{\mathrm{d} (SA)}{\mathrm{d} (DFT)} \cdot \frac{\mathrm{d} (DFT)}{\mathrm{d} x} 
\end{equation}

For backward, the critical part is the second term $\frac{\mathrm{d} (Norm)}{\mathrm{d} (SA)}$. The general form of normalization function is $y=\frac{t_1}{C}$, where $C=\sum_j^N{t_j}$. The derivative of $y$ over $t$ is computed as follow:

\begin{equation}\label{eq11}
	D(t) = 
	\begin{cases}
	      \frac{\partial y_{i} }{\partial t_{i}} =\frac{ct_{i}^{'}-t_{i}t_{i}^{'} }{C^{2} } , & if \ i=j \vspace{2ex}\\
	      \frac{\partial y_{i} }{\partial t_{i}} =-\frac{t_{i}t_{i}^{'} }{C^{2} } , & if \ i\neq j
	\end{cases}
\end{equation}

For Softmax function, $t_i$=$e^{x_i}$, $C_1=\sum{e^{x_j}}$, the derivative over $x$ is:
\begin{equation}\label{eq12}
	D(x) = 
	\begin{cases}
	       \frac{\partial y_{i} }{\partial x_{j} } =\frac{e^{x_{i} }C_{1}-(e^{x_{i} } )^{2}   }{{C_{1}}^{2} }  , & if \ i=j \vspace{2ex}\\
	       \frac{\partial y_{i} }{\partial x_{j} } =-\frac{e^{x_{i} }e^{x_{j} }}{{C_{1}}^{2} }  , & if \ i\neq j
	\end{cases}
\end{equation}

In Fourier complex field, the scale of $x_i$ is extremely larger than the spatial field. Hence $e^{x_i}e^{x_j}$ and ${C_1}^{2}$ in Eq. \eqref{eq12} are unstable and easy to be out of range. As to $Logmax$ function, $t_i$=$log|{x_i}|$, $C_2=\sum{log|{x_j}|}$. Hence the derivative can be calculated as below:
\begin{equation}\label{eq13}
	D(x) = 
	\begin{cases}
	       \frac{\partial y_{i} }{\partial x_{j} } =\frac{C_{2}-\log_{}{|x_{i}| }  }{|x_{i}|{C_{2}}^{2} }   , & if \ i=j \vspace{2ex}\\
	       \frac{\partial y_{i} }{\partial x_{j} } =-\frac{\log_{}{|x_{i}| }  }{|x_{i}|{C_{2}}^{2} }   , & if \ i\neq j
	\end{cases}
\end{equation}

Rather than $C_{1}$ in Eq. \eqref{eq12}, $C_{2}$ in Eq. \eqref{eq13} is smoothed by logarithm operation, which is critical for stabilizing the gradient backward when training SA in Fourier complex field.

\subsubsection{Complex self-attention}With $Logmax$ function, we investigate the architecture of gathering contextual information on the Fourier complex field and further propose a simple but effective model named Fourier Complex Transformer (shown in Fig. \ref{fig_architecture} (a)). In each FCT block, the input $x$ of size $[H \times W \times C]$ is processed by the first LayerNorm block. Then the size of $x$ is flattened into $[C \times HW]$ and is fed into the proposed Complex self-attention unit to calculate the contextual relationship. In CSA, the global representation $X\in\mathbb{R}^{C \times \frac{HW}{2}}$ is obtained through DFT(x). After that, $Q, K, V$ is projected from $X$ by three 1D convolution layers $q(x), k(x), v(x)$. Next, we separate the complex values $Q, K$ and $V$ into their real symmetrical parts ${Q_r,K_r, V_r}$ and imaginary anti-symmetrical parts ${Q_i,K_i, V_i}$. As shown in Fig. \ref{fig_csa}, the attention maps $Attn_r$ and $Attn_i \in\mathbb{R}^{\frac{HW}{2} \times \frac{HW}{2}}$ are computed via a modified SA, which replace the $Softmax$ function with the proposed $Logmax$ function:
\begin{equation}\label{eq14}
    Attn=Logmax(Q^{T} K)
\end{equation}

Then we use Eqs. \eqref{eq15}-\eqref{eq16} to calculate the fused contextual representation, where $\alpha$ is a learnable scale position embedding and "$\cdot$" denotes the dot product.
\begin{align}
    \label{eq15}
    CSA_{r} =[\alpha \cdot Attn_{r}+(1-\alpha )\cdot Attn_{i}]V_{r}^{T}  \\ 
    \label{eq16}
    CSA_{i} =[\alpha \cdot Attn_{i}+(1-\alpha )\cdot Attn_{r}]V_{i}^{T}
\end{align}

Compared with the naive additional position embedding, the learnable scale position embedding adaptively scales the attention map and exchanges information between the real and imaginary attention maps, achieving better fused contextual representation. After that, the results of $CSA_r$ and $CSA_i$ are grouped as one complex value and transformed back into the real spatial field by IDFT. After CSA, the second LayerNorm and MLP are appended to model the local bias. Finally, we use GELU as the activation function. As depicted in Fig. \ref{fig_architecture}b, there are two extra residual connections in each FCT block.
\begin{figure*}[b]
    \centering
    \includegraphics[width=0.9\linewidth]{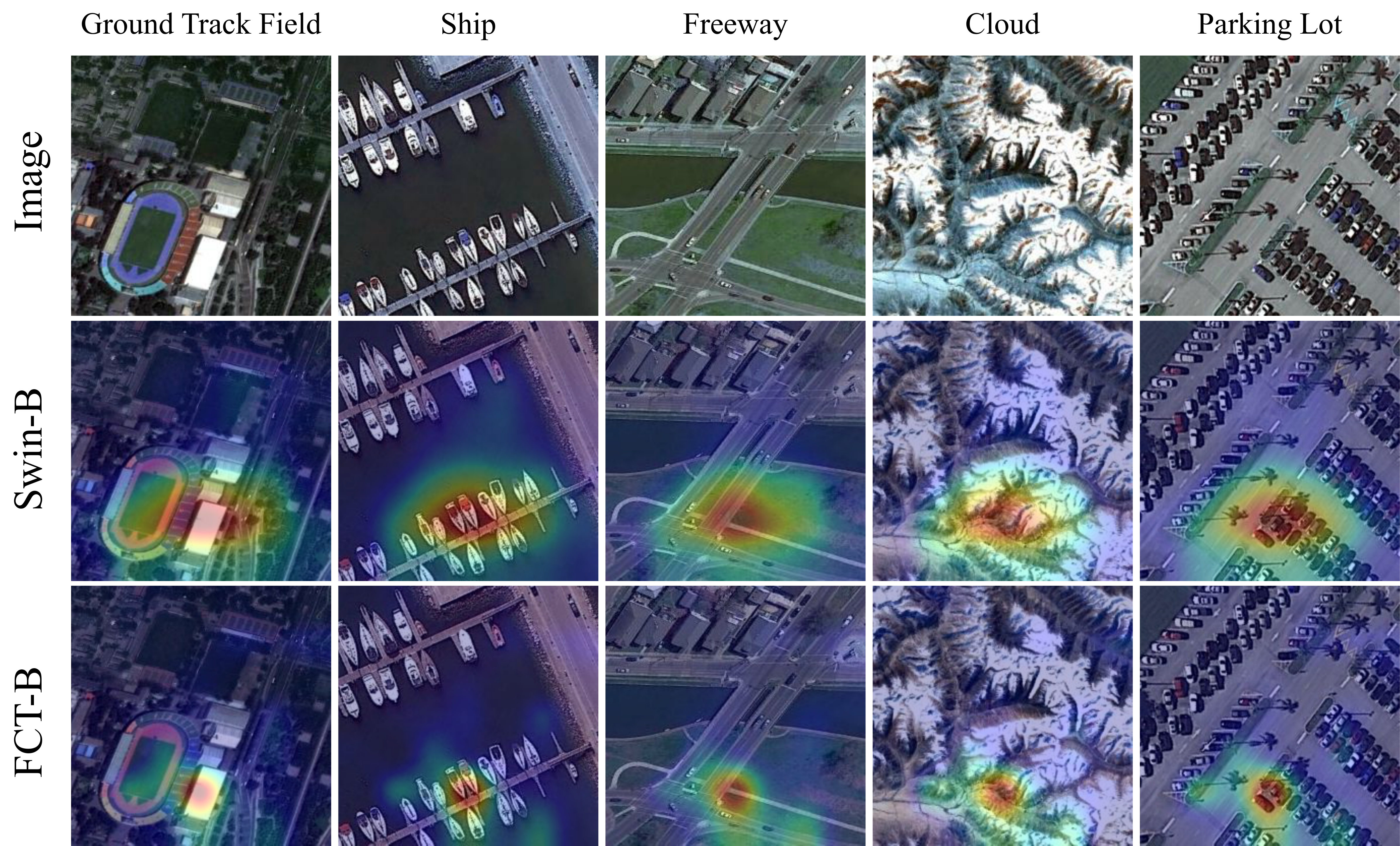}
    \caption{Some samples of class activate map, key area visualizations of Swin-B (2nd row) and FCT-B (3nd row).}
    \label{fig_cam_of_FCT}
\end{figure*}

\subsection{Efficiency Analysis and Architecture Design}
In this section, we give a theoretical analysis of the efficiency of the proposed FCT. In FCT block, the complexity of DFT, IDFT, and CSA is $nlog(n), nlogn$ and $\frac{1}{2}O(n^2)$, respectively, $n$ is the size of patches. Hence the total complexity of CSA is about $2nlogn+1/2O(n^2)$, which is much less than the naive SA ($O(n^2)$) while $n>>log(n)$. 

From the analysis above, the proposed FCT block is much more efficient in processing large-scale feature maps. Moreover, it is possible to further be improved by incorporating CSA with WSA proposed in Swin-transformer, which is suitable for VHR RS imagery. By stacking FCT blocks, the proposed FCT is designed as a hierarchical architecture, where $HW>>C$ in the first 2 stages, and $C>>HW$ in the last 2 stages. Hence, we use the spatial-FCT block in the first 2 stages, which performs CSA on the spatial dimension $(H, W)$ to learn the abundant spatial contextual relationship. In the last 2 stages, We use a modified channel-FCT, which learns the channel contextual information via CSA to learn the contextual relationships on the channel dimension $(C)$. Following the common design of Transformer-based model\cite{vit,swin}, we propose four versions of FCT with various architecture hyper-parameters:
\begin{itemize}
    \item FCT-Tiny: $C_1$ = 96, layer numbers = $\left\{3, 3, 6, 3\right\}$
    \item FCT-Small: $C_1$ = 96, layer numbers = $\left\{3, 6, 12, 3\right\}$
    \item FCT-Base: $C_1$ = 128, layer numbers = $\left\{3, 6, 12, 3\right\}$
    \item FCT-Large: $C_1$ = 192, layer numbers = $\left\{3, 6, 12, 3\right\}$
\end{itemize}

\begin{table}[h]
\caption{Parameters and FLOPs of different FCT Architecture.}
    \centering
    \begingroup
    \setlength{\tabcolsep}{5pt} 
    \renewcommand{\arraystretch}{1} 
    
    \begin{tabular}{l  c  c  c  c }
    \toprule
    Architecture & FCT-T & FCT-S & FCT-B & FCT-L   \\ \midrule \midrule
    Image Size & 224 & 224 & 224 & 224  \\
    Parameters(M) & 27.4 & 52.6 & 74 & 179 \\
    FLOPs(G) & 4.9 & 8.9 & 15.3	& 38 \\ \midrule
    
    Image Size	& 512 & 512	& 512 & 512	\\
    Parameters(M) & 28.2 & 54.8 & 78.2	& 174.5 \\
    FLOPs(G) & 25.7 & 42.4	& 69.8 & 169.2 \\ \bottomrule

    \end{tabular}
    \label{tab_architecture}
    \endgroup
\end{table}

where $C_1$ is the channel number of the hidden layers in the ﬁrst stage, and the channel is doubled in each later stage. Table \ref{tab_architecture} compares the parameters and theoretical computational complexity (FLOPs) of FCT under various architectures, revealing the efficiency of the proposed FCT for VHR RS images.

\section{Experiments and Analysis}
\subsection{Scene Classification}

\subsubsection{Data sets}We conduct experiments on 3 widely used single-label scene classification data sets. The AID data set\cite{aid} has number of 10000 images within 30 aerial scene types, by collecting sample images from Google Earth imagery and a spatial resolution ranging from 0.5 to 0.8m. Each class of the data sets contains 220–420 images with a size of 600$\times$600 pixels.
\begin{table}[t]
\caption{Comparison OA($\%$) results on AID and NWPU data set. For AID and NWPU-45, training rates are set to be 20\%, 50\% and 10\%, 20\%, respectively. }
    \centering
    \begingroup
    \setlength{\tabcolsep}{5pt} 
    \renewcommand{\arraystretch}{1} 
    
    \begin{tabular}{ l  c  c  c  c }
    \toprule
    \multirow{2}*{Network} & \multicolumn{2}{c}{AID} & \multicolumn{2}{c}{NWPU-45} \\
    \cmidrule(r){2-3} \cmidrule(r){4-5} &T.R.=20$\%$ &T.R.=50$\%$ & T.R.=10$\%$ &T.R.=20$\%$  \\ \midrule \midrule
    SCCov\cite{sccov}	&93.12	&96.10	&89.30	&92.10 \\
    MG-CAP\cite{mgcap}	&93.34	&96.12	&90.83	&92.95 \\
    GRMA-Net\cite{grma} &94.55	&96.98	&91.56	&93.22 \\
    Attn-2\cite{attn2}	&95.37	&96.56	&-	    &- \\
    CAD\cite{cad}	    &95.73	&97.16	&92.70	&94.58 \\
    KFBNet\cite{kfbnet}	&95.50	&97.40	&93.08	&95.11 \\
    MGML\cite{mgml}	&96.45	&98.60	&92.91	&95.39 \\
    ResNet-50\cite{resnet} &93.56 &96.69	&91.26	&94.61 \\
    ResNext-50\cite{resnext} &93.61 &96.74 &91.56	&94.87 \\
    GFNet\cite{gfnet}	&94.36 &97.24 &91.75 &94.44 \\
    ViT-L\cite{vit}   &94.05 &97.01	&90.55	&94.33 \\
    Swin-T\cite{swin}	&94.56 &97.41   &92.28	&94.97 \\
    Swin-B	&95.69	&98.20	&92.33	&95.28 \\
    Swin-L	&96.33	&\textbf{98.90}	&92.61	&95.76 \\
    \textbf{FCT-T}	&95.93	&98.56	&92.12	&95.45 \\
    \textbf{FCT-S}	&96.06	&98.62	&92.28	&95.84 \\
    \textbf{FCT-B}	&96.57	&98.65	&92.31	&95.85 \\
    \textbf{FCT-L}	&\textbf{96.89}	&98.87	&\textbf{93.01}	&\textbf{96.03} \\\bottomrule
    \end{tabular}
    \endgroup
    \label{tab_aid}
\end{table}
All the aerial images of AID are carefully selected from various countries and regions all over the world, mainly in China, USA, UK, France, Japan, Germany, etc., and they are obtained at varied times and seasons under various imaging conditions, which results in significant intra-class diversities of the data. 

The NWPU-RESISC45 data set\cite{nwpu} is a large-scale remote sensing image data set for image scene classification, which contains 31,500 images and 45 scene classes. Each class consists of 700 images with a size of 256$\times$256 pixels. For most scene classes, NWPU-RESISC45 contains varying spatial resolutions ranging from about 30 to 0.2 m per pixel. 

The ERA single image data set\cite{era} is a recently proposed data set for event recognition in remote sensing images/videos and is taken by UAVs, which consists of 2,864 images/videos each with a label from 25 different classes. In this paper, we only use images with a spatial size of 640$\times$640 pixels, which is much more challenging than AID and NWPU data set.

\begin{figure*}[b]
    \centering
    \includegraphics[width=0.8\linewidth]{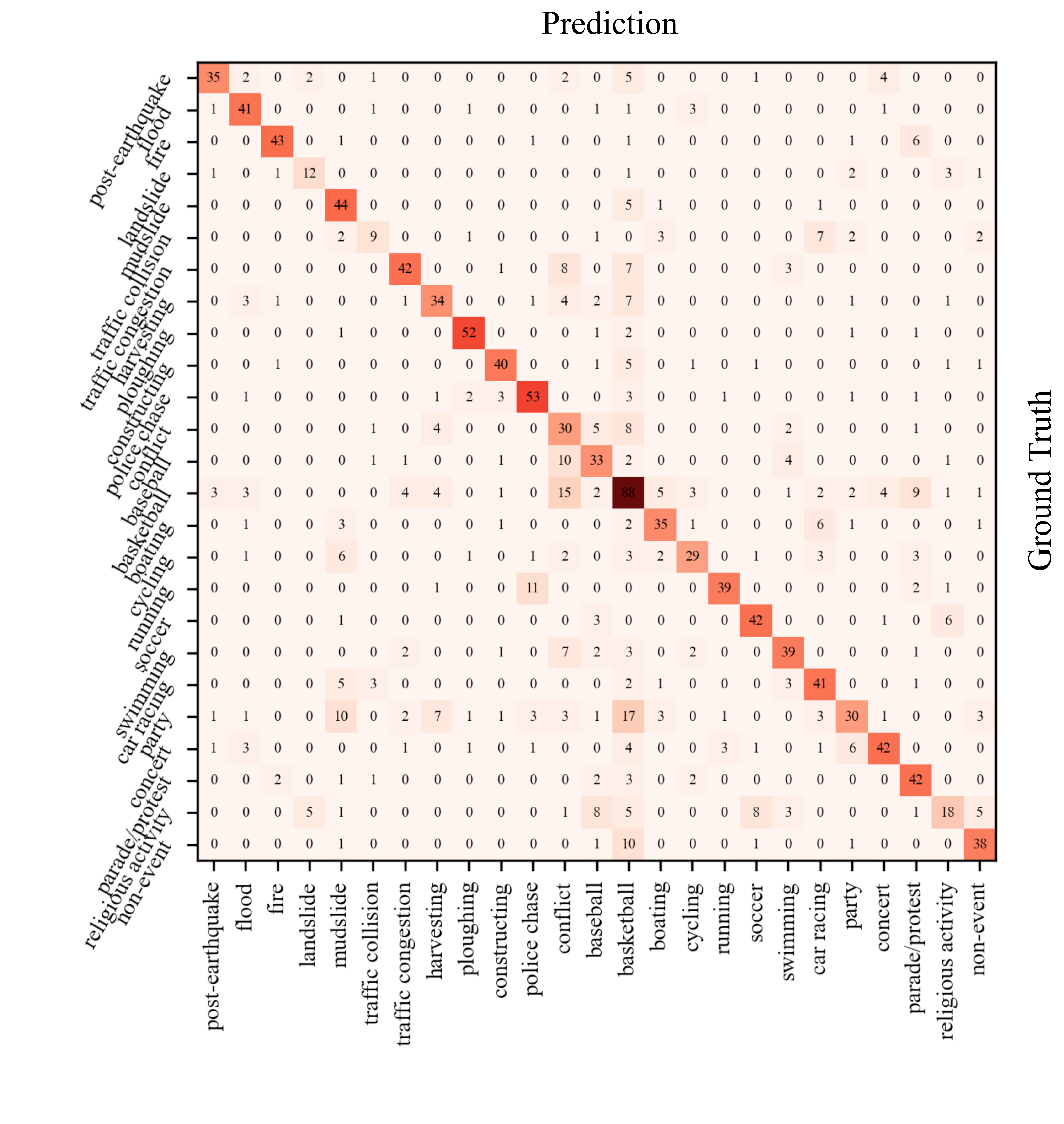}
    \caption{Confusion Matrix of the Era data set using the proposed FCT-B model.}
    \label{fig_confusing_matrix}
\end{figure*}

\begin{table}[t]
\caption{Comparison results on Era data set. Parameter and FLOPs are measured using the GitHub repository of MMclassification\cite{mmclassification} with image resolution 600$\times$600.}
    \centering
    \begingroup
    \setlength{\tabcolsep}{5pt} 
    \renewcommand{\arraystretch}{1} 

    \begin{tabular}{ l  c  c  c }
    \toprule
    Network & OA ($\%$) & Params(M) & FLOPs(G)  \\ \midrule \midrule
    VGG-16\cite{vgg} & 51.91 &134.3 &15.5 \\
    VGG-19 & 49.77 &143.7 &19.6 \\
    ResNet-50\cite{resnet} & 57.32 &23.6 &43.0 \\
    ResNet-101 & 55.34 &42.5 &86.8 \\
    ResNet-152 & 56.10 &58.2 &121.0 \\
    MobileNet\cite{mobilenetv2} & 61.30 &5.48 &1.18 \\
    DenseNet-121\cite{dense} & 61.75 &8.9 &33.0 \\
    DenseNet-169 & 60.67 &13.9 &37.7 \\
    DenseNet-201 & 62.33 &32.7 &45.8 \\
    GFNet\cite{gfnet} & 63.12 & 86 &137.0 \\
    Swin-B\cite{swin} & 64.60 & 88.9 &87.5 \\
    \textbf{FCT-B} & \textbf{66.41} & 86.8 &70.3 \\\bottomrule
    \end{tabular}
    \endgroup
    \label{tab_era}
\end{table}
\subsubsection{Experimental setting and metric}Following most of the related work, we use AdamW as the optimizer, the learning rate is set to 0.001, and the other hyperparameter of AdamW are set as default in Pytorch. We use the polygon schedule to adjust the learning rate. During training scene classification model, we resize the image into 256$\times$256 with the commonly used random rotation and crop augment operations. Image with the original size is used for the test. For semantic segmentation, we simply use random rotation during training. All the models of scene classification are initialized with the parameter pre-trained on ImageNet and conducted on Nvidia RTX 3090 with a batch size of 64. We apply the overall accuracy (OA) to measure the performance of our model, which is defined as:
\begin{equation}\label{eq17}
    OA=\frac{\sum_{k=1}^{N} TP_{k} }{N}
\end{equation}

Where $TP_k$, is the $k$-th correctly classified test sample, $N$ is the total number of the test samples.

\subsubsection{Performance comparison}To demonstrate the overall single-image classification performance, we mainly evaluate the proposed FCT on two RS scene classification data sets: AID and NWPU-45. Another more challenging ERA data set for event recognition is also used for extra evaluation. Following previous work, we randomly select 20$\%$/50$\%$ of AID and 10$\%$/20$\%$ of NWPU as the train set, and the rest are remained as the test set. For a fair comparison, we not only compare the proposed FCT with general image classification networks like ResNet, ViT, PVT\cite{pvt} and Swin-transformer, but also involve some specifically designed methods, i,e. SCCovnet\cite{sccov}, MGML\cite{mgml}, MG-CAP\cite{mgcap}, GRMA-Net\cite{grma}, Attn-2\cite{attn2}, CAD\cite{cad} and KFBNet\cite{kfbnet} for VHR RS image scene classification task. The results are listed in Table \ref{tab_aid}.

For the AID data set, several specifically designed networks achieve remarkable results on both 20$\%$ and 50$\%$ training ratio. Because previous work apply various networks as the backbone, we just report the best results in Table \ref{tab_aid}. With 20$\%$ training ratio, MGML got 98.6$\%$ OA on AID data set due to an extra block designed to enhance local information. Rather than these specially designed methods, general backbones achieve comparable results as well. Some very recent Transformer-based backbone models surpass CNN-based backbones in a huge margin with comparable parameters. The currently popular Swin-T achieves 94.56$\%$ and 97.41$\%$ on 20$\%$ and 50$\%$ train ratio and therefore confirms the necessity to model long-range contextual relationship. Besides the aforementioned model, the efficient version of the proposed FCT named FCT-T surpasses Swin-T by 1.37$\%$ and 1.55$\%$ when T.R is 20$\%$ and 50$\%$, respectively. When we use FCT-L, 96.89$\%$ and 98.87$\%$ OA can be achieved, which is to-date the best-published performance on the AID data set in the default setup. 

Similar results are observed in the NWPU data set. FCT-L achieves 93.01$\%$ OA on 10$\%$ training ratio, just 0.05$\%$ less than last SOTA KFBNet. When increasing training ratio into into 20$\%$, FCT-L obtains the new SOTA 96.03$\%$, which is 0.27$\%$ higher than Swin-L and is the unique method with OA higher than 96$\%$. Additionally, we can find from Table \ref{tab_aid} that Transformer-based methods (ViT, Swin and FCT) achieve better performance while T.R.=20$\%$. This phenomenon denotes that rather than the CNN-based model, the Transformer-based models are more data hungry either in real spatial field or Fourier complex field. Some class activate maps (CAM) are visualized in Fig. \ref{fig_cam_of_FCT}. These maps show some representative samples that reflect the reasonable impact of key area location. Compared with Swin-transformer, the FCT model can pay more attention to the key area location, which is helpful to improve classification accuracy.

As shown in Table \ref{tab_era}, unified performance degeneration is observed on the very challenging ERA data set due to the less inter-class variance. From Table \ref{tab_era}, the performance of many common-used CNN-based backbones are reported. Among them, DenseNet-201 achieves the best result on 62.3$\%$ OA. Within a very close parameter scale, the overall accuracy of our FCT-B can achieve 66.4$\%$, making the absolute improvement over the best competitor GFNet and Swin-B by 3.3$\%$ and 1.8$\%$, respectively. The results on the ERA data set demonstrate the great ability of the proposed FCT on learning discriminative representation from large resolution images (640$\times$640) with complex texture.
The confusing matrix of FCT for the ERA data set is presented in Fig. \ref{fig_confusing_matrix}. Most of the misclassification samples belong to the classes of landslide, traffic collision, conflict, baseball, cycling, party, and religious activity. It is mainly because there are lots of confusing objects and features among these scene classes that limit the classification performance.

\subsection{Object Detection}
\begin{table*}[t]
\caption{Comparison results of class-wise AP on DOTA-v1.0. RN and CMR indicates RetinaNet and Cascade Mask R-CNN, respectively. The short names are defined as Baseball diamond (BD), Ground field track (GTF), Small vehicle (SV), Large vehicle (LV), Tennis court (TC), Basketball court (BC), Storage tank (ST), Soccer-ball field (SBF), Roundabout (RA), Swimming pool (SP), Helicopter (HC).}
    \centering
    \begingroup
    \setlength{\tabcolsep}{3.8pt} 
    \renewcommand{\arraystretch}{1} 

    \begin{tabular}{l  c  c  c  c  c  c  c  c c c c c c c c c c}
    \toprule
    
    Method & Backbone & Plane & BD & Bridge & GTF & SV & LV & Ship & TC & BC & ST & SBF & RA & Harbor & SP & HC & mAP \\ \midrule \midrule
    \multirow{1}[9]{*}{RN\cite{RetinaNet}}      & Res-101\cite{resnet} & 86.54 & 77.45 & 42.8  & 64.87 & 71.06 & 58.5  & 73.53 & 90.72 & 80.97 & \textbf{66.67} & 52.42 & 62.16 & 60.79 & 64.84 & 40.83 & 66.28 \\
    & GFNet-S\cite{gfnet} & 87.32 & 78.48 & 43.25 & 65.42 & 71.64 & 58.63 & 73.98 & 90.08 & 79.28 & 65.14 & 52.96 & 64.87 & 62.7  & 66.14 & 42.38 & 66.82 \\
    & Swin-S\cite{swin} & 88.65 & 78.16 & 43.5  & 65.44 & 71.69 & \textbf{59.26} & 74.38 & 90.69 & 81.36 & 66.48 & 52.94 & \textbf{65.14} & 62.44 & 66.09 & 43.1  & 67.29 \\
    & \textbf{FCT-S} & \textbf{90.1} & \textbf{79.04} & \textbf{43.68} & \textbf{65.9} & \textbf{71.58} & 57.98 & \textbf{75.16} & \textbf{90.9} & \textbf{81.73} & 66.39 & \textbf{53.57} & 65.06 & \textbf{63.19} & \textbf{66.28} & \textbf{43.35} & \textbf{67.59} \\ \midrule
    \multirow{2}[7]{*}{CMR\cite{cascade}} & Res-101\cite{resnet} & 88.93 & 75.21 & 51.55 & 64.9  & 74.39 & 75.37 & 84.74 & 90.23 & 77.48 & 81.51 & 46.57 & 63.49 & 65.39 & 67.63 & 56.96 & 70.96 \\
    & GFNet-S\cite{gfnet} & 89.51 & 76.33 & 51.69 & 65.37 & 74.66 & 75.81 & 86.15 & 90.67 & 78.96 & 81.14 & 48.14 & 65.71 & 66.68 & 69.38 & 58.12 & 71.89 \\
    & Swin-S\cite{swin} & \textbf{89.58} & 75.89 & 51.78 & 65.15 & 74.85 & 75.87 & 85.48 & 90.58 & 78.54 & 82.67 & \textbf{48.67} & 64.99 & 66.24 & 68.43 & 58.69 & 71.83 \\
    & \textbf{FCT-S} & 89.49 & \textbf{76.97} & \textbf{51.97} & \textbf{65.88} & \textbf{74.89} & \textbf{76.03} & \textbf{86.72} & \textbf{90.78} & \textbf{80.16} & \textbf{82.73} & 48.54 & \textbf{65.79} & \textbf{67.4} & \textbf{69.69} & \textbf{58.77} & \textbf{72.39} \\
    
    \bottomrule
    \end{tabular}
    \endgroup
    \label{tab_ob}
\end{table*}

\subsubsection{Data sets}In this section, we use the popular DOTA data set\cite{dota} to test the effectiveness of FCT on RS image object detection task. DOTA data set contains 1,793,658 instances from 2806 high-resolution aerial images (4000$\times$4000) of 18 typical RS categories. Each image is sliced into 25 800$\times$800 pieces. Different from the general image object detection data set, annotations in the DOTA data set include both the horizontal boundary box (HBB) and the orientation horizontal boundary (OBB). All object detection results reported in this part is tested on OBB.

\subsubsection{Experimental setting and metric}For a fair comparison, we consider the typical 1-stage RetinaNet (RN)\cite{RetinaNet} and 2-stage Cascade Mask R-CNN (CMR)\cite{cascade} as the baseline frameworks. Following most of the work \cite{object1,object2,object3}, we utilize multi-scale training (longer side is at most 1200), 3x schedule (36 epochs with learning rate decayed at epochs 24, 30 by the factor of 0.1) and AdamW optimizer with default hyper-parameter in MMDetection\cite{mmdetection}. All the models are initialized with the parameter pre-trained on ImageNet-22K. We use the mean average precision (mAP) to measure the performance of object detection models. The detailed computation of mAP can be referred to \cite{voc}.

\begin{figure*}[h]
    \centering
    \includegraphics[width=\linewidth]{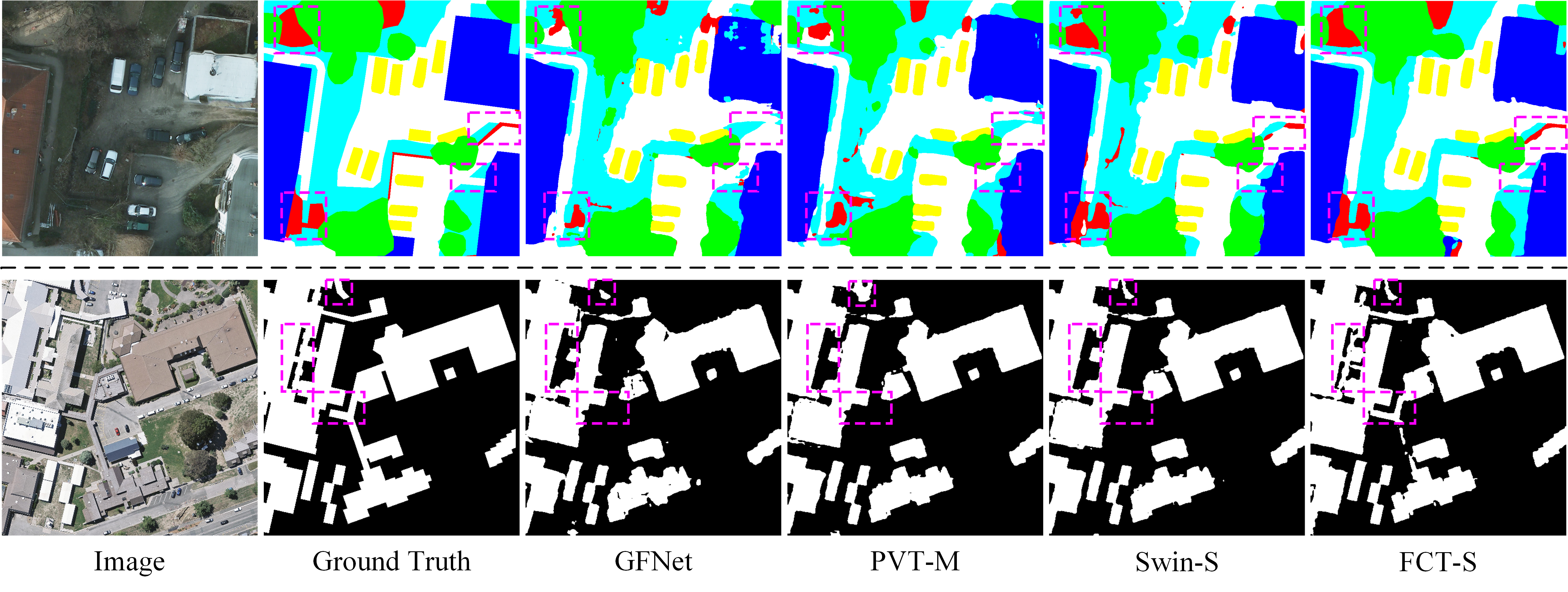}
    \caption{Mapping results on ISPRS Potsdam data set (top). Impervious surface: white, building: dark blue, low vegetation: light blue, tree: green, car: yellow and background: red. Mapping results on WHU data set (bottom) Building: white, clutter: black.}
    \label{fig_segmentation_mapping_result}
\end{figure*}
\subsubsection{Performance comparison}The results of RS image object detection task are listed in Table \ref{tab_ob}. With the typical 1-stage framework of RetinaNet\cite{RetinaNet}, naive ResNet-101\cite{resnet} achieves 66.28 mAP. By replacing ResNet with latest SOTA backbones, GFNet\cite{gfnet} and Swin-transformer\cite{swin} gain stable improvements (0.54 mAP and 1.01 mAP) over ResNet. Rather than the above-mentioned methods, RetinaNet with the backbone of the proposed FCT achieves 67.59 mAP, surpassing the previous best results by 0.3 mAP. For the 2-stage CMR\cite{cascade} method, very similar performance boosts can be observed. Our FCT achieves 72.39 mAP, which is 0.56 mAP and 0.5 mAP higher than  Swin-transformer and GFNet, respectively. Also, we found that FCT gains remarkable results on large-scale objects (Baseball diamond, basketball court, roundabout and Harbor), which mainly benefited from the global high-order feature representation.

\begin{table*}[h]
\caption{Comparison results on Potsdam data set.}
    \centering
    \begingroup
    \setlength{\tabcolsep}{5.7pt} 
    \renewcommand{\arraystretch}{1} 

    \begin{tabular}{l  c  c  c  c  c  c  c  c c c c}
    \toprule
    \multirow{2}*{Method} &  \multirow{2}*{Backbone} &\multicolumn{6}{c}{IoU($\%$)} & \multirow{2}*{mIoU($\%$)} & \multirow{2}*{Pacc($\%$)} & \multirow{2}*{Params(M)} & \multirow{2}*{FLOPs(G)} \\
    
    \cmidrule{3-8}& & Imp. surf. & Building &  Low veg. &  Tree & Car & Clutter\\ \midrule \midrule
    \multirow{3}[10]{*}{DeepLabV3+\cite{deeplabv3+}} & Res-101\cite{resnet} & 86.63 & 92.59 & 75.81 & 78.60  & 91.22 & 41.10  & 77.66 & 90.22 &254.6 &62.6 \\
    & ViT-B\cite{vit} & 86.03 & 91.27 & 76.47 & 78.41 & 90.55 & 40.13 & 77.14 & 90.50 &61.2 &91.43\\
    & PVT-M\cite{pvt}   & 86.42 & 92.76 & 76.69 & 78.97 & 90.18 & 37.97 & 77.16 & 90.32 &53.3 & 66.2\\
    & GFNet\cite{gfnet}  & 87.04 & 93.22 & 75.66  & 78.61 & \textbf{92.78} & 39.15  & 77.76  & 90.55 &61.2 &73.8 \\
    & Swin-S\cite{swin} & 86.95 & 92.90  & 77.95 & 79.45 & 90.40  & \textbf{44.51} & 78.69 & 90.87 &51.4  &53.8\\
    & \textbf{FCT-S} & \textbf{88.21} & \textbf{93.90} & \textbf{78.86} & \textbf{80.33} & 92.17  & 42.50 & \textbf{79.33} & \textbf{91.60}  &49.6  & 52.3\\
    \midrule
    \multirow{3}[8]{*}{SegFormer\cite{segformer}} & Res-101\cite{resnet} & 86.89 & 93.41 & 75.95 & 78.83 & 92.14 & 40.68 & 77.98 & 90.45 &179.2 &42.5\\
    & ViT-B\cite{vit} & 86.99 & 93.26 & 75.37 & 78.94 & 91.72 & 38.15 & 77.47 & 90.33 &63.1 &69.6\\
    & PVT-M\cite{pvt} & 86.69 & 93.20 & 75.77 & 78.66 & 92.80 & 39.11 & 77.70 & 90.43 &63.1 &69.6\\ 
     & GFNet\cite{gfnet} & 87.11 & 93.79 & 77.12 & 79.93 & 91.78 & 41.02 & 78.45 & 90.50 &51.9 &81.9 \\
    & Swin-S\cite{swin} & 87.13 & 93.47 & 77.33 & 79.92 & 92.02 & \textbf{44.60} & 79.07 & 91.25 &56.8 &49.8\\
    & \textbf{FCT-S} & \textbf{88.54} & \textbf{94.70} & \textbf{79.31} & \textbf{81.23} & \textbf{93.68}  & 42.06 & \textbf{79.92} & \textbf{92.10} &52.6 &49.3 \\
    \bottomrule
    \end{tabular}
    \endgroup
    \label{tab_isprs}
\end{table*}

\begin{table}[h]
\caption{Comparison results on WHU data set.}
    \centering
    \begingroup
    \setlength{\tabcolsep}{3pt} 
    \renewcommand{\arraystretch}{1} 

    \begin{tabular}{l c  c  c  c  c  c}
    \toprule
    \multirow{2}*{Method} &  \multirow{2}*{Backbone} &\multicolumn{2}{c}{IoU($\%$)} & \multirow{2}*{mIoU($\%$)} & \multirow{2}*{Pacc($\%$)}  \\
    \cmidrule{3-4}& & Building & Background \\ \midrule \midrule
    \multirow{3}[10]{*}{DeepLabV3+}
    & Res-101\cite{resnet}  & 87.74 & 98.30  & 93.02 & 98.51 \\
         & ViT-B\cite{vit}  & 86.94 & 97.46 & 92.20 & 98.37  \\
         & PVT-M\cite{pvt}  &88.33 &98.23 &93.28  & 98.60 \\
        & GFNet\cite{gfnet}  & 89.12 & 98.89 & 94.00 & 98.76 \\
       & Swin-S\cite{swin} & 87.95  & 98.49 & 93.17 & 98.67  \\
        & \textbf{FCT-S} &  \textbf{89.80}  &  \textbf{98.80} &  \textbf{94.35}  & \textbf{98.82} \\
    \midrule
    \multirow{3}[8]{*}{SegFormer}
    & Res-101\cite{resnet} & 87.74 & 97.70 & 92.72 & 98.28  \\
         & ViT-B\cite{vit} & 87.89 & 98.41 & 93.10 & 98.54 \\
         & PVT-M\cite{pvt}  &87.29 & 98.45 &92.87 &98.30  \\
         & GFNet\cite{gfnet}  & 89.43 & 98.39 & 93.86 & 98.76 \\
         & Swin-S\cite{swin} & 89.71 & 98.49 & 94.10 & 98.80 \\
         & \textbf{FCT-S} &  \textbf{90.56} &  \textbf{98.74}  &  \textbf{94.65}    & \textbf{98.91} \\
    \bottomrule
    \end{tabular}
    \endgroup
    \label{tab_whu}
\end{table}

\subsection{Semantic Segmentation}
\subsubsection{Data sets}Two widely used remote sensing semantic segmentation data sets are used in this section. The ISPRS Potsdam data set\cite{isprs} is a challenging remote sensing scene-parsing data set containing 38 fine spatial resolution images of size 6000$\times$6000 pixels divided into 6 common land cover categories with a ground sampling distance (GSD) of 5 cm. The data set provides near-infrared, RGB as well as DSM and normalized DSM (NDSM). We employed only the RGB images in the experiments. Following the official suggestion, we select 24 images as the train set and the remaining 14 images as the test set. Considering that the images are too large to be used as input, we cropped the training images into 500$\times$500 patches without overlap.

The WHU data set\cite{whu} contains 2 categories with about 22,000 independent buildings. The original remote sensing data comes from the New Zealand Land Information Services website. The ready-to-use samples consist of 8,189 images with 512$\times$512 pixels.

\subsubsection{Experimental setting and metric}Semantic segmentation is a typical kind of pixel-wise classification task. Different from image-level classification, pixel-wise classification has a higher requirement for local context modeling. Therefore, we use the proposed FF-SA as the encoder of many widely used semantic segmentation models to test its contextual modeling ability.

For a fair comparison, all experiments are conducted on the MMSegmentation\cite{mmseg} platform under the same setting. All the training processes were implemented on NVIDIA GeForce RTX 3090 with a batch size of 32. AdamW is used as the optimizer. We set the initializing learning rate and weight decay to 3×$10^{-4}$, and 0.01, respectively. We simply use random rotation during training to avoid over-fitting. All models were trained with 80k iterations for convergence.

We use Pixel Accuracy (Pacc) and mIoU as the metrics to measure the segmentation performances of different models. The computation of PAcc is shown as Eq.\eqref{eq18}. Pacc is an accurate metric, but scores of different segmentation model on Pacc is not apparent. Therefore, we use the mIoU, which is computed as Eq.\eqref{eq19} as the supplement metric.
\begin{align}
    \label{eq18}
    Pacc=\frac{\sum_{k=1}^{N} T P_{k}}{\sum_{k=1}^{N} T P_{k}+T N_{k}+F P_{k}+F N_{k}} \\
    \label{eq19}
    mIoU=\frac{1}{N} \sum_{k=1}^{N} \frac{T P_{k}}{T P_{k}+F P_{k}+F N_{k}}
\end{align}

Where $TP_k$, $TN_k$, $FP_k$, and $FN_k$ indicate the true positive, true negative, false positive, and false negatives, respectively, for object indexed as class $k$. The mIoU can better measure the segmentation performance on difficult areas like boundaries and small objects.

\subsubsection{Performance comparison}Different from image-level classification, pixel-wise classification has a higher requirement for local context modeling. Therefore, we compare FCT with the latest popular backbones on 2 widely used semantic segmentation baselines (DeepLabV3+\cite{deeplabv3+}, Segformer\cite{segformer}) to test the local contextual modeling ability.

We first evaluate the comparison results on ISPRS Potsdam Data set, which are listed in Table \ref{tab_isprs}. All segmentation frameworks achieve 76$\%$ higher mIoU and 90$\%$ higher Pacc with proper backbones. Specifically, the comparative backbone not only include the classical and efficient models i.e., ResNet, ViT, Swin-transformer, and PVT\cite{pvt}, but also consider the Fourier transform models i.e., GFNet. The numeric scores for the test data set illustrated that FCT generates high accuracy, surpassing other competitive backbones including both the CNN-based models and the Transformer-based models in the Pacc and mIoU by an obvious margin.

The detailed segmentation accuracy on the WHU data set of each network is listed in Table \ref{tab_whu}. All the methods achieve better performance on the background and poorer performance on the buildings. The buildings are difficult to recognize because the buildings are easily confused with other things, such as roads and cars in the background. Rather than DeeplabV3+, SegFormer works better with most backbones and shows great adaptation with a transformer-based backbone. The proposed SegFormer-FCT-S model achieves 94.6$\%$ mIoU, which not only achieves the best performance but also has significant domination on FLOPs and Parameters amount.

\subsection{Ablation Study}
In this section, we separately ablate the critical design elements in the proposed FCT with the challenging NWPU remote sensing image scene classification data set.
\begin{table}[h]
\caption{Ablation study of each component in FCT. s, c indicate spatial-FCT block and channel-FCT block. }
    \centering
    \begingroup
    \setlength{\tabcolsep}{4pt} 
    \renewcommand{\arraystretch}{1} 

    \begin{tabular}{c  c  c  c  c  c }
    \toprule
    Architecture & APE  & SPE & Logmax & OA($\%$) & FPS    \\ \midrule \midrule
     & \checkmark  &   &   & 94.77 &751 \\
    \textbf{s, s, c, c} &    & \checkmark &   & 95.34 & 753 \\
     &   & \checkmark & \checkmark	& \textbf{95.84}	& 714 \\ \midrule
     & \checkmark  &  	&   & 94.79	& 758 \\
    \textbf{s, s, s, c} &    & \checkmark	&   & 95.43	& 757 \\
     &   &  \checkmark & \checkmark & 95.81 & 706 \\  \midrule
     & \checkmark  &  	&   & 94.30	& 864 \\
    \textbf{s, s, s, s} &    & \checkmark &  	& 94.62 & 865 \\
     &   & \checkmark	& \checkmark & 95.04 & 770 \\ \bottomrule 

    \end{tabular}
    \endgroup
    \label{tab_ablation}
\end{table}

\subsubsection{Replacement Study}Firstly, we investigate the influence of components in the proposed FCT, including architecture variance, position embedding, and normalized function. The results are listed in Table \ref{tab_ablation}, where $s, c$, APE, and SPE represent spatial-FCT, channel-FCT, naive additional position embedding, and learnable scale attention embedding, respectively. Among three FCT architectures, $(s, s, c, c)$ achieves the best trade-off between performance and source costs. Rather than APE, SPE is much more suitable for embedding position information on the Fourier complex and gains general improvements in all settings. Results in Table \ref{tab_ablation} further prove the superiority of the proposed \textit{Logmax} on improving the classification accuracy.

\subsubsection{Effect of \textit{Logmax} Function}
\begin{figure}[t]
    \centering
    \includegraphics[width=\columnwidth]{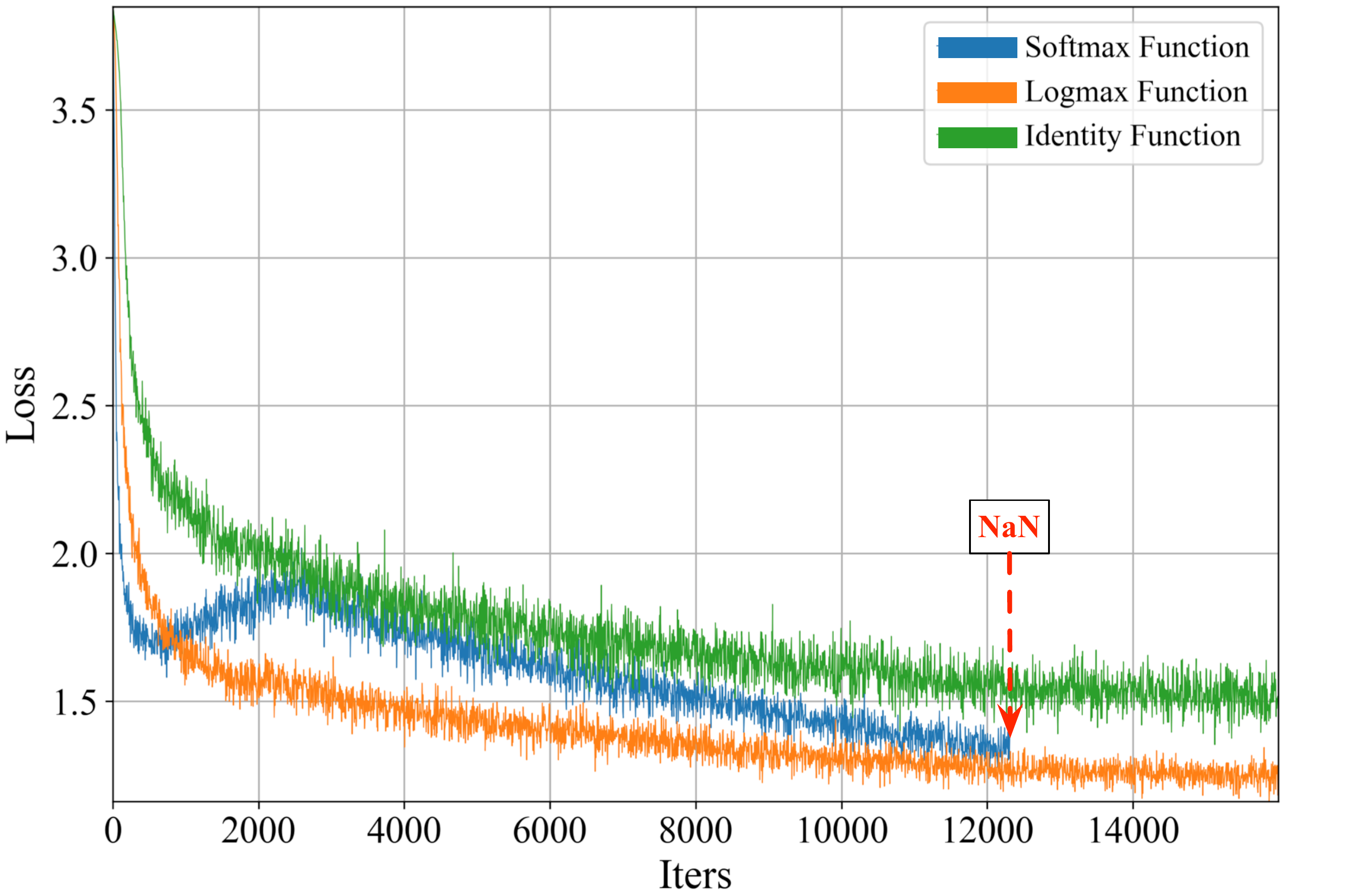}
    \caption{Loss curve of various normalized function. \textit{Logmax} is converged with lower loss value, while the loss of FCT with \textit{Softmax} function is crashed into Nan near 12000th iteration.}
    \label{fig_lossfuction}
\end{figure}

Secondly, we evaluate the effectiveness of \textit{Logmax} function   in FCT, the loss curves of FCT with \textit{Softmax} function, \textit{Logmax} function and identity function are shown in Fig. \ref{fig_lossfuction}. With identity function $(y=x)$, loss is unstable during training and converged in a local minimum. \textit{Softmax} function has better performance but suffered from gradient unstable. Rather than the above 2 functions, the proposed \textit{Logmax} function achieves the smoothest and lowest loss value, which promises the effectiveness of \textit{Logmax} function on normalizing tokens in the Fourier complex field.

\subsubsection{Parameters, FLOPs vs Image Resolution}
\begin{figure}[t]
    \centering
    \includegraphics[width=\columnwidth]{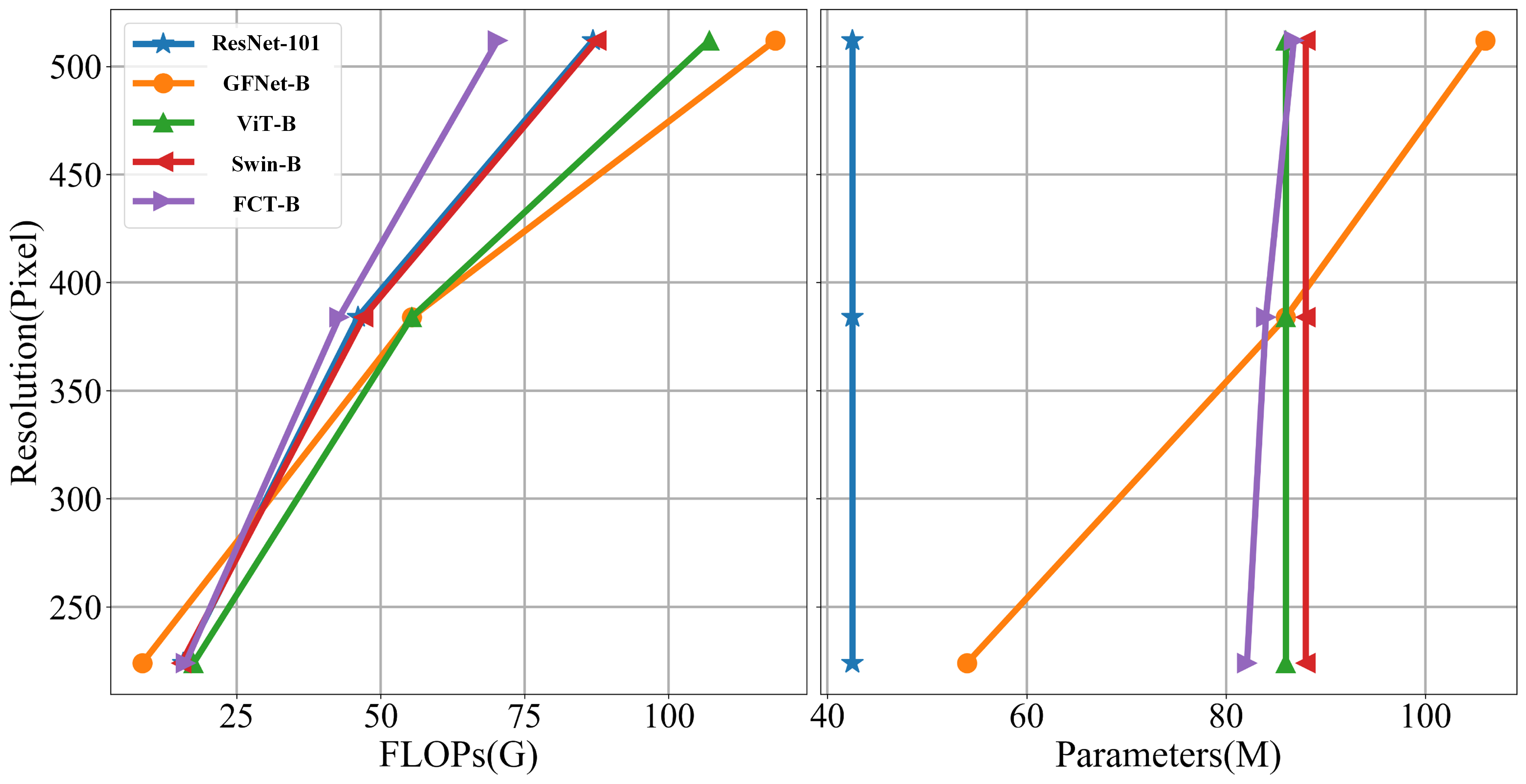}
    \caption{FLOPs-resolution (left) and parameters-resolution (right) curves.}
    \label{fig_FLOPs}
\end{figure}

Here we evaluate our FCT and other novel models regarding both memory, computational costs, and image resolution for scene classification tasks. Fig. \ref{fig_FLOPs} shows the FLOPs-Resolution and Parameters-Resolution curves under default training setup, respectively for FCT and many popular vision backbones. As shown in Fig. \ref{fig_FLOPs}, the proposed FCT consistently obtains better memory and computation efficiency over all image resolution. Especially on images with a size larger than 300, FCT gains definite advantages on memory and computation costs. This character makes FCT suitable for processing VHR RS images.

\begin{table}[h]
\caption{Ablation study of different position embedding methods on three benchmarks, using the FCT-T architecture. abs. pos.: absolute position embedding term of ViT\cite{vit}; rel. pos.: relative position bias term of Swin\cite{swin}; scl. pos.: scale position embedding term (see Eqs. \eqref{eq15}-\eqref{eq16}).}
    \centering
    \begingroup
    \setlength{\tabcolsep}{8pt} 
    \renewcommand{\arraystretch}{1} 

    \begin{tabular}{l c  c  c}
    \toprule
    \multirow{2}*{ } &  \makecell[c]{AID\\\hline Acc} & \makecell[c]{DOTA\\ \hline mAP}  & \makecell[c]{Potsdam\\ \hline mIoU} \\ \midrule \midrule
    
    no pos.             &94.01  &78.12  &63.45     \\
    abs. pos.           &94.45  &78.44  &63.21     \\
    rel. pos.           &95.15  &79.02  &66.97     \\
    abs. + rel. pos.      &95.07  &78.95  &66.89     \\
    \textbf{scl. pos. (our)}    &\textbf{95.93}  &79.33  &\textbf{67.59}     \\
    scl. + abs. pos.      &95.91  &79.15  &67.31     \\
    scl. + rel. pos.      &\textbf{95.93}  &\textbf{79.41}  &67.55     \\
    scl. + abs. + rel. pos. &95.89  &79.32  &67.57     \\
    \bottomrule
    \end{tabular}
    \endgroup
    \label{tab_position}
\end{table}
\subsubsection{Scale Position Bias}Table \ref{tab_position} shows comparisons of different position embedding approaches. FCT-T with the proposed scale position embedding yields +1.48$\%$/0.78$\%$ accuracy on the AID data set, +0.89/0.31 mAP on the DOTA data set, +4.38$\%$/0.62$\%$ mIoU on Potsdam data set in relationship with those absolute position embedding and relative position bias, respectively, which demonstrates the effectiveness of scale position embedding. When the relative position bias is included, the performance of object detection is improved (+0.08mAP on DOTA), but the performance of semantic segmentation is impaired (-0.04$\%$ mIoU on Potsdam). We also note that it undermines the performances on the 3 tasks while all of the 3 position embedding methods are integrated. This phenomenon supports our assumption that the absolute position embedding breaks the information position consistency in Fourier complex field.

\subsubsection{Generalization on Natural Image}To test the generalization of the proposed FCT, we evaluate our FCT on regular ImageNet-1K data set in Table \ref{tab_imagenetcompared}. Compared with the currently popular transformer-based methods, such as DeiT\cite{deit} and Swin-transformer\cite{swin}, FCT achieve superior results with similar computational costs: +1.1$\%$/1.7$\%$ for FCT-T/B over DeiT-S/B using 224$^2$ input. With higher image resolution, most of the methods obtain better results. Among them, the proposed FCT achieves the best trade-off between performance and efficiency. FCT-B with 512$^2$ images significantly outperforms other methods within real-time computational costs.

\begin{table}[t]
\caption{Comparison of different backbones on ImageNet-1K classification. Throughput is measured following Swin\cite{swin}.}
    \centering
    \begingroup
    \setlength{\tabcolsep}{3pt} 
    \renewcommand{\arraystretch}{1} 
    
    \begin{tabular}{l  c  c  c c c}
    \toprule
    \multicolumn{6}{c}{\textbf{Regular ImageNet-1K trained models}} \\
    Method & \makecell[c]{image size\\(pixel)} &\makecell[c]{Params\\(M)} &\makecell[c]{FLOPs\\(G)}& \makecell[c]{throughput\\(image/s)} & \makecell[c]{ImageNet\\top-1 acc.}   \\ \midrule \midrule
    RegNetY-4G\cite{regnety} & 224$^2$ & 21 & 4.0 &1156.7 &80.0  \\
    RegNetY-8G & 224$^2$ & 39 & 8.0 &591.6 &81.7  \\
    RegNetY-16G & 224$^2$ & 84 & 16.0 &334.7 &82.9  \\
    EfficientNet-B3\cite{efficientnet} & 300$^2$ & 12 & 1.8 &732.1 &81.6  \\
    EfficientNet-B4 & 380$^2$ & 19 & 4.2 &349.4 &82.9  \\
    EfficientNet-B5 & 456$^2$ & 30 & 9.9 &169.1 &83.6  \\
    EfficientNet-B6 & 528$^2$ & 43 & 19.0 &96.9 &84.0  \\
    EfficientNet-B7 & 600$^2$ & 66 & 37.0 &55.1 &84.3  \\
    ViT-B/16\cite{vit} & 384$^2$	& 86 & 55.4	&85.9 &77.9  \\
    ViT-L/16 & 384$^2$	& 307 &190.7 &27.3	&76.5 \\
    DeiT-S\cite{deit} & 224$^2$ &22 &4.6 & 940.4 & 79.8 \\
    DeiT-B &224$^2$ &86 &17.5 &292.3 &81.8 \\
    DeiT-B &384$^2$ &86 &55.4 &85.9 &83.1 \\
    Swin-T\cite{swin} &224$^2$ &29  &4.5 &755.2 &81.3 \\
    Swin-S &224$^2$ &50  &8.7 &436.9 &83.0 \\
    Swin-B &224$^2$ &88  &15.4 &278.1 &83.3 \\
    Swin-B &384$^2$ &88  &47.0 &84.7 &84.2 \\
    \textbf{FCT-T} &224$^2$ &27 &4.9 &732.0 &80.9 \\
    \textbf{FCT-S}&224$^2$ &52 &8.9 &438.8 &83.1 \\
    \textbf{FCT-B} &224$^2$ &78 &15.3 &280.1 &83.5 \\
    \textbf{FCT-B} &384$^2$ &78 &40.8 &100.0 &84.6 \\
    \textbf{FCT-B} &512$^2$ &78 &69.8 &59.8 &\textbf{85.0} \\
    \bottomrule
    

    \end{tabular}
    \label{tab_imagenetcompared}
    \endgroup
\end{table}

\section{Conclusion}
In this paper, we have presented the Fourier Complex Transformer (FCT) to tackle the VHR RS image classification task. In FCT, a novel Complex self-attention (CSA) is proposed to efficiently learn the contextual relationships in the Fourier complex field. We design the \textit{Logmax} function to stabilize gradient in back propagation. Extensive experiments not only show the effectiveness of FCT on the VHR RS image scene classification and segmentation benchmark but also support the distinctive efficiency superiority of FCT in processing high-resolution images, which is essential for building efficient down-stream model for VHR RS images. In the near future, we will further explore the interpretation of the contextual relationships, thus making the classification results more reliable. The computational complexity of FCT is not refined, which is also worth studying in the future.

\bibliographystyle{IEEEtran}
\bibliography{references}

\end{document}